\documentclass[10pt,twocolumn,letterpaper]{article}

\usepackage{iccv}
\usepackage{times}
\usepackage{epsfig}
\usepackage{amsmath}
\usepackage{amssymb}

\usepackage{algorithm}
\usepackage{algpseudocode}

\usepackage{indentfirst}
\usepackage{longtable}
\usepackage{subfigure}
\usepackage{multirow}
\usepackage{graphicx}
\usepackage{makecell}
\usepackage{url}
\def\Figref#1{Figure~\ref{#1}}
\def\Tabref#1{Table~\ref{#1}}
\def\Secref#1{Section~\ref{#1}}
\def\Algref#1{Algorithm~\ref{#1}}
\def\Eqref#1{{Eq.~\eqref{#1}}}
\def\RMw{\mathrm{UniformSample}}
\def\RMo{\mathrm{OrderedUniformSample}}

\def\NAME{{SETN}}
\usepackage{amsmath,amsfonts,bm}

\def\va{{\bm{a}}}

\def\vf{{\bm{f}}}
\def\vg{{\bm{g}}}
\def\vh{{\bm{h}}}

\def\mH{{\bm{H}}}
\def\mI{{\bm{I}}}

\def\gA{{\mathcal{A}}}

\def\gD{{\mathcal{D}}}

\def\gI{{\mathcal{I}}}

\def\gO{{\mathcal{O}}}

\def\gR{{\mathcal{R}}}

\def\gT{{\mathcal{T}}}

\def\sA{{\mathbb{A}}}

\def\sW{{\mathbb{W}}}

\usepackage{etoolbox}
\makeatletter
\patchcmd{\maketitle}
 {\def\@makefnmark}
 {\def\@makefnmark{}\def\useless@macro}
 {}{}
\makeatother

\usepackage[pagebackref=true,breaklinks=true,,colorlinks,bookmarks=false]{hyperref}

\iccvfinalcopy % *** Uncomment this line for the final submission

\def\iccvPaperID{0126} % *** Enter the ICCV Paper ID here

\ificcvfinal\fi

\begin{document}

%%%%%%%%% TITLE
%\title{Quality-Aware One-Shot Neural Architecture Search}
%\title{Performance-Aware One-Shot Neural Architecture Search}
%\title{Performance-Aware Template Network for One-Shot Neural Architecture Search}
\title{One-Shot Neural Architecture Search via Self-Evaluated Template Network}

\author{
Xuanyi Dong$^{\dagger\ddagger}$\thanks{This paper was accepted to the IEEE ICCV 2019.}\thanks{This work was done when Xuanyi Dong was a research intern with Baidu Research.} and Yi Yang$^{\ddagger}$ \\
%$^{\dagger}$University of Technology Sydney, $^{\ddagger}$Baidu Research\\
$^{\dagger}$Baidu Research, $^{\ddagger}$ReLER, University of Technology Sydney\\
{\tt\small xuanyi.dxy@gmail.com}, {\tt\small yi.yang@uts.edu.au}
}

\maketitle
\ificcvfinal\thispagestyle{empty}\fi

\begin{abstract}
Neural architecture search (NAS) aims to automate the search procedure of architecture instead of manual design. Even if recent NAS approaches finish the search within days, lengthy training is still required for a specific architecture candidate to get the parameters for its accurate evaluation. Recently one-shot NAS methods are proposed to largely squeeze the tedious training process by sharing parameters across candidates. In this way, the parameters for each candidate can be directly extracted from the shared parameters instead of training them from scratch. However, they have no sense of which candidate will perform better until evaluation so that the candidates to evaluate are randomly sampled and the top-1 candidate is considered the best. In this paper, we propose a Self-Evaluated Template Network ({\NAME}) to improve the quality of the architecture candidates for evaluation so that it is more likely to cover competitive candidates. {\NAME} consists of two components: (1) an evaluator, which learns to indicate the probability of each individual architecture being likely to have a lower validation loss. The candidates for evaluation can thus be selectively sampled according to this evaluator. (2) a template network, which shares parameters among all candidates to amortize the training cost of generated candidates. In experiments, the architecture found by {\NAME} achieves the state-of-the-art performance on CIFAR and ImageNet benchmarks within comparable computation costs. Code is publicly available on GitHub: \url{https://github.com/D-X-Y/AutoDL-Projects}.
\end{abstract}

%%%%%%%%% Introduction
\section{Introduction}

Representation learning~\cite{bengio2013representation} is a fundamental research problem in computer vision, because it is beneficial to a variety of computer vision applications, such as detection and segmentation.
Due to the success of deep learning, it has undergone a transition from ``feature engineering''~\cite{lowe1999object,dalal2005histograms} to ``architecture engineering''~\cite{szegedy2015going,huang2017densely,simonyan2015very,liu2019ppn,he2016deep,dong2017more}. However, a large amount of expert knowledge and ample computational resources are still required to secure an architecture for good feature representations~\cite{zoph2017NAS}.
Fortunately, neural architecture search (NAS) brings hope to deep learning researchers and alleviate their labours~\cite{Zoph_2018_CVPR,pmlr-v80-pham18a}.
% Fortunately, automated architectures, i.e., neural architecture search (NAS), brings hope to deep learning researchers and alleviate their labours~\cite{Zoph_2018_CVPR,pmlr-v80-pham18a}.
%Recently, NAS has attracted more and more attention and became an important problem in the computer vision community.

The goal of NAS is to discover an optimal network in the search space, which can maximize the validation accuracy after training. Typical algorithms apply reinforcement learning (RL)~\cite{zoph2017NAS,Zoph_2018_CVPR,pmlr-v80-pham18a,chen2019renas} or evolutionary strategy~(EA)~\cite{liu2018hierarchical,real2019regularized} to solve this problem, and most are computationally expensive, e.g., 500 GPUs over four days~\cite{Zoph_2018_CVPR}. Such huge computational costs motivate the researchers to focus on efficient architecture search algorithms. Recently, several works reduced the computational cost through weight sharing~\cite{pmlr-v80-pham18a,cai2018efficient}, weight generation~\cite{brock2018smash,bender2018understanding}, accuracy prediction~\cite{baker2018accelerating,klein2017learning}, progressive strategy~\cite{Liu_2018_ECCV}, etc.\looseness-1

\begin{figure}[t!]
\begin{center}
\includegraphics[width=\linewidth]{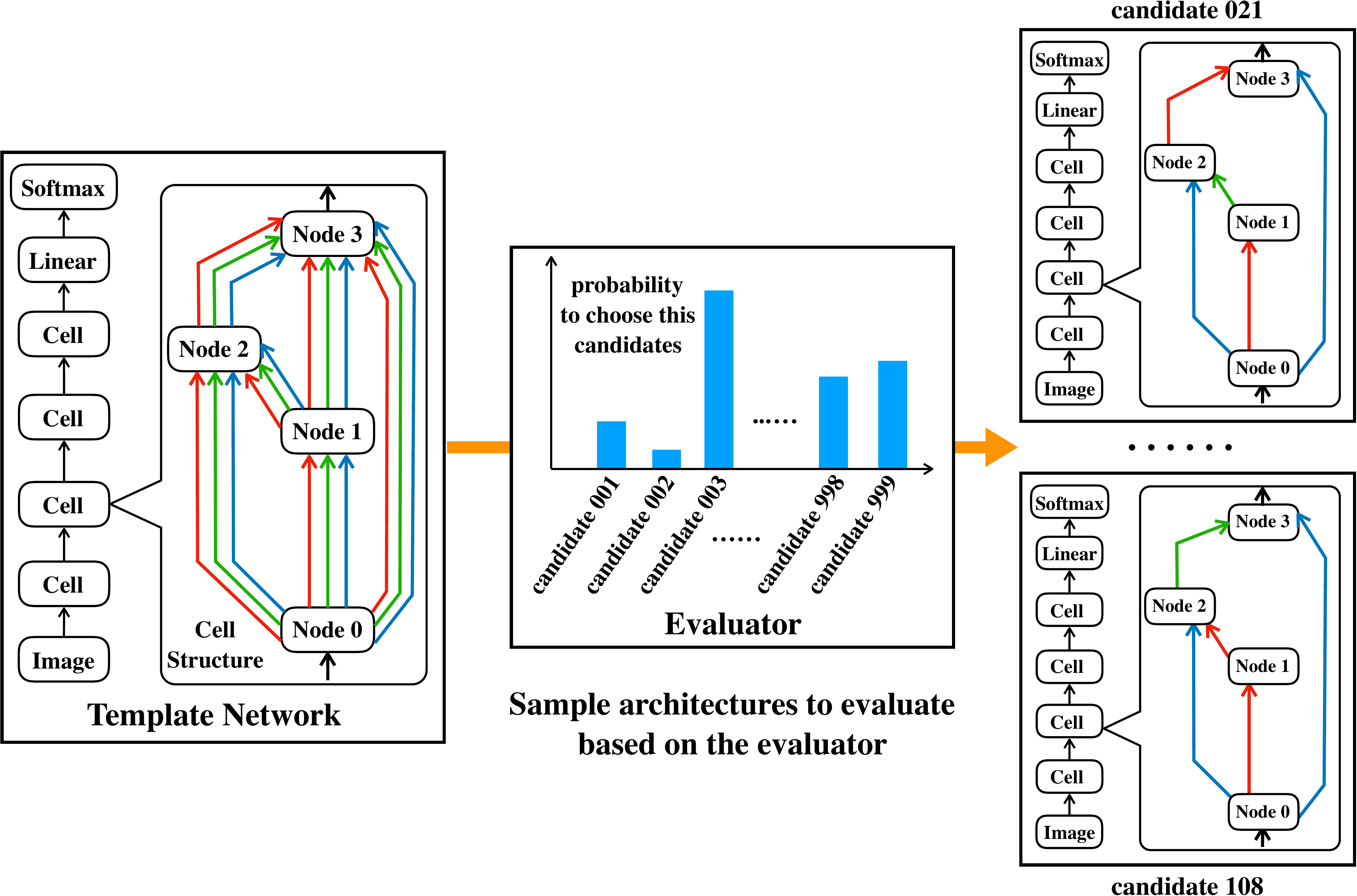}
\end{center}
\vspace{-3mm}
\caption{
An overview of the self-evaluated template network ({\NAME}). 
{\NAME} consists of a template network and an evaluator.
Architectures for evaluation are generated by sampling candidates in the template network using the evaluator.
The template network shares the parameters for different candidates as indicated by connections of arrow lines in different colors. The evaluator learns the distribution of the architectures being likely to have a lower validation loss.
}
\vspace{-2mm}
\label{fig:intro}
\end{figure}

One-shot NAS approaches stand out among these efficient NAS approaches~\cite{bender2018understanding,brock2018smash}, because they can significantly squeeze the tedious training process by sharing parameters across architecture candidates.
A typical one-shot NAS paradigm is to (1) randomly sample hundreds of architecture candidates {from the parameter-shared network}; (2) evaluate these candidates; and (3) find the candidate with the highest validation accuracy.
We observed that most of these randomly sampled candidates are useless, since most of them have a poor performance. Besides, these sampled candidates are only a small portion of the whole search space; therefore {the probability for inclusion of the best architecture is extremely low}.
% the best architecture could not even be selected, and the performance of the discovered architecture is limited.
%Some of them can directly obtain parameters of a CNN without a time-consuming training procedure.
%They randomly select thousands of candidate CNNs with the generated parameters and find one CNN with the highest validation accuracy in one or two GPU days~\cite{brock2018smash,bender2018understanding}.
%Unfortunately, there are billions of different networks in the search space~\cite{Liu_2018_ECCV}, and the randomly selected networks are just a very small portion.
%Therefore, the best network could even be not selected by these algorithms, and the performance of the discovered network is restricted.

To solve the above problem, we propose \textit{Self-Evaluated Template Network ({\NAME})} for one-shot NAS.
{\NAME} equips a template network with an evaluator.
The \textit{template network} contains all possible candidate convolutional neural networks (CNNs) and shares its parameters (template parameters) with all candidates. We train this template network in \textit{a stochastic strategy}: during one iteration, we uniformly sample one candidate and only optimize partial template parameters for this sampled candidate. In this way, after training, each candidate CNN can directly use the corresponding template parameters without additional training.
Some previous methods coupled the shared parameters with a learnable distribution of architectures, such as~\cite{liu2019darts,dong2019search,pmlr-v80-pham18a}.
These methods will introduce the bias into the template parameters.
Since some shallow and light-weight network will quickly converged, the learnable distribution will bias to these ``simple'' networks, and other networks may not have a chance to be updated.
In contrast to them, our uniformly stochastic training strategy allows each candidate to be treated equally. In this way, each candidate with shared template parameters would be trained fully, and their validation accuracy would be closer to the ground truth validation accuracy.

The \textit{evaluator} learns to indicate the probability of each individual candidate CNN being likely to have a lower validation loss. We train this evaluator on the validation data with the assistance of the template network.
After training, according to the learned probability, our evaluator can pick up low-validation-loss candidates for one-shot evaluation.
As a result, the probability of including the best CNN in the search space can be dramatically improved.
Compared to previous random sampling algorithms~\cite{bender2018understanding,brock2018smash}, {\NAME} can potentially search for a CNN with a higher accuracy.

In experiments, {\NAME} can discover a superior CNN on CIFAR-10 within two GPU days.
This {\NAME}-searched architecture achieves state-of-the-art performance on three benchmarks, i.e., CIFAR-10, CIFAR-100, and ImageNet.

\section{Related Work}\label{sec:related-work}

Automatically discovering effective networks has attracted more and more researchers~\cite{brock2018smash,baker2017designing,zoph2017NAS,liu2018hierarchical,fang2020densely,dong2019network}. Various kinds of searching algorithms have been proposed, such as RL-based~\cite{zoph2017NAS,Zoph_2018_CVPR,pmlr-v80-pham18a}, EA-based~\cite{real2019regularized,real2017large}, gradient-based~\cite{liu2019darts,luo2018neural} approaches. Each of these algorithm has its unique advantage.
For simplicity, we summarize some ubiquitous algorithms in \Tabref{table:compare-method-attribute} regarding five aspects.
Our {\NAME} has advantages when compared to them, and we will introduce these related works at below. Note that we focus on searching CNN models, and thus approaches about recurrent neural network or downstream applications are out of the scope of this paper.

\begin{table}[t!]
\centering
\setlength{\tabcolsep}{2.6pt}
\begin{tabular}{| c | c | c | c | c | c |}\hline
                                         & Search    &   Eff.    & Share    & Eval.  & Gen. \\
\hline
  ResNet~\cite{he2016deep}               & Manual    &  $\times$ & $-$      & $-$    & $-$    \\
\hline
  MetaQNN~\cite{baker2017designing}      & RL        &  $\times$ & $\times$ & $-$    & $-$    \\
  NASNet~\cite{Zoph_2018_CVPR}           & RL        &  $\times$ & $\times$ & $-$    & $-$    \\
  AmoebaNet~\cite{real2019regularized}   & EA        &  $\times$ & $\times$ & $-$    & $-$    \\
   H-NAS~\cite{liu2018hierarchical}      & EA        &  $\times$ & $\times$ & $-$    & $-$    \\
  EAS~\cite{cai2018efficient}            & RL        &  $\surd$  & $\surd$  & $-$    & $-$    \\
  PNAS~\cite{Liu_2018_ECCV}              & SMBO      &  $\times$ & $\times$ & slow   & $-$    \\
  SMASH~\cite{brock2018smash}            & Gradient  &  $\surd$  & $\times$ & quick  & random \\
Understand~\cite{bender2018understanding}& Gradient  &  $\surd$  & $\surd$  & quick  & random \\
  DARTS~\cite{liu2019darts}              & Gradient  &  $\surd$  & $\surd$  & $-$    & $-$    \\
  GDAS~\cite{dong2019search}             & Gradient  &  $\surd$  & $\surd$  & $-$    & $-$    \\\hline
  Our {\NAME}                            & Gradient  &  $\surd$  & $\surd$  & quick  & selective  \\
\hline
\end{tabular}
\vspace{1mm}
\caption{
We compare different algorithms with five aspects.
``Search'' shows the search algorithm type.
``Eff.'' indicates efficient, and we consider the algorithm that can discover CNN within five GPU days as efficient.
``Share'' indicates whether the algorithm shares parameters over different candidate networks or not.
``Eval.'' means whether the algorithm is able to quickly evaluate a network.
``Gen.'' indicates the network sampling strategy during \textit{the one-shot evaluation procedure}.
``-'' indicates not available, which means those methods do not have these steps.
}
\vspace{-4mm}
\label{table:compare-method-attribute}
\end{table}

Early approaches train a large amount of candidate networks by tens epochs and use the validation accuracy of these networks as the supervisory signal~\cite{liu2018hierarchical,real2019regularized,zoph2017NAS,Zoph_2018_CVPR,real2017large}.
For example, Zoph~et~al.~\cite{zoph2017NAS,Zoph_2018_CVPR} learn an RL policy to sample networks with high accuracy.
Real~et~al.~\cite{real2017large,real2019regularized} utilize EA algorithms to mutate low-quality networks to high-quality networks.
Unfortunately, these approaches cost too much computational resources.
Recent approaches aim to solve the searching problem in an affordable computation cost.
Liu~et~al.~\cite{Liu_2018_ECCV} progressively search CNN from simple to complex. Barker~et~al.~\cite{baker2018accelerating} accelerate the searching procedure by performance prediction. Pham~et~al.~\cite{pmlr-v80-pham18a} share the parameters of different networks.
Liu~et~al.~\cite{liu2019darts} propose DARTS, allowing efficient search of the architecture using gradient descent.
DARTS~\cite{liu2019darts} and GDAS~\cite{dong2019search} choose the best architecture using the $\arg\max$ over a continues architecture representation, while the performance of an architecture cannot be correctly estimated without fully training it.
Our {\NAME} can more accurately estimate the performance (validation loss) of all candidates without separately training each of them one by one. 
%Different from the above efficient approaches, our SAS trains a single TemplateNet, and then each candidate CNN can reuse the template parameters without the tediousness of training each of them one by one.
%In this way, we avoid the heavy training load from a different perspective and can efficiently discover a robust CNN in experiments.

Our {\NAME} is one-shot NAS, and is closely related to previous one-shot approaches \cite{zhang2019graph,bender2018understanding,brock2018smash}.
Brock~et~al.~\cite{brock2018smash} train hypernetworks~\cite{ha2017hypernetworks} to generate suitable weights for every network in the search space. Zhang~et~al.~\cite{zhang2019graph} encode the network as a computation graph and use a graph neural network to predict weights.
Bender~et~al.~\cite{bender2018understanding} deliver thorough experimental analysis for one-shot architecture search.
These approaches can estimate the network performance correctly without additional training, while the architecture candidates for evaluation are randomly picked with uneven quality~\cite{bender2018understanding,brock2018smash}.
Our approach can selectively sample low-validation-loss architecture candidates.
In this way, our sampled candidates would have much lower validation loss (better performance) than that of the random strategy~\cite{bender2018understanding,brock2018smash}, and thus we can potentially discover a more effective CNN architecture.
%
% Randomly sample hundreds of candidates and rank these networks according to the validation accuracy.
% In contrast, we selectively generate high-quality candidate CNN instead of randomly generation.
% In this way, we can dramatically reduce the search space and potentially discover more robust CNNs.

\section{Background}\label{sec:background}

Early works search for the whole CNN structure~\cite{zoph2017NAS,bello2017neural}, whereas recent works propose that finding a good neural cell is more effective than finding a whole CNN~\cite{Liu_2018_ECCV,liu2019darts,zhang2019graph,dong2019search}.
Therefore, we also search for a good cell instead of a full CNN model.
%and thus $\alpha$ represents the neural cells in our work.
As shown in \Figref{fig:intro}, a cell is a fully convolutional structure, mapping a tensor $\mI_{in} \in \gR^{H\times W \times C}$ to another tensor $\mI_{out} \in \gR^{H'\times W'\times C'}$.
If we use the stride of 1, the cell is named normal cell, which has $(H', W', F')$=$(H, {W}, {F})$; and if we use the stride of 2, the cell is named reduction cell, which has $({H'}, {W'}, {F'})$=$(\frac{H}{2}, \frac{W}{2}, {2F})$.
Each cell contains $B$ nodes, where each of them is specified as a quadruple $(\mI_{1},\mI_{2},f_{1},f_{2})$~\cite{Liu_2018_ECCV}.
Specifically, the $i$-th node in the $c$-the cell takes two inputs $\mI_{1},\mI_{2}\in \gI_{i}^{c}$ and generates a tensor $\mH_{i}^{c}=f_{1}(\mI_{1}) + f_{2}(\mI_{2})$.
$f_{1},f_{2} \in \gO$ are transformation functions to apply to inputs.
The output of the $c$-th cell is the concatenation of each intermediate output tensor from each node, denoted as $\mH^{c}$.

The set of possible inputs $\gI_{i}^{c}$ is the output set of all previous nodes adding the outputs of two previous cells: {$\gI_{i}^{c} = \{ \mH_{1}^{c}, ..., \mH_{i-1}^{c}, \mH^{c-1}. \mH^{c-2}\}$}. The candidate function set $\gO$ contains several pre-defined functions.
In this paper, we apply our {\NAME} on the candidate function set $\gO$ following previous methods~\cite{liu2019darts,dong2019search} as follows:
\begin{table}[H]
\small
\vspace{-3mm}
\setlength{\tabcolsep}{0.3em} % for the horizontal padding
\centering
\def\arraystretch{1.0}
\begin{tabular}{lll}
$\bullet$ 3x3 max pooling     & $\bullet$ 3x3 avg pooling  & $\bullet$ skip connection     \\
$\bullet$ 3x3 separable conv  & $\bullet$ 5x5 separable conv & $\bullet$ 1x3 \& 3x1 conv \\
\end{tabular}
\vspace{-3mm}
\end{table}
\noindent We set the number of nodes in a cell as $B=4$.
Therefore, the number of candidates in the search space of $\gO$ is $(1\times3\times6\times10\times{(6^2)}^{4})^{2} = 9.1\times10^{16}$.
% ; and we also try {\NAME} on a large candidate function set $\gO_{l}$ to evaluate our efficiency.
% The small function set $\gO_{s}$ contains six different functions as:
% 
% \noindent The large function set $\gO_{l}$ extends $\gO_{s}$ with two more functions, having eight functions in total:
% \begin{table}[H]
% \small
% \setlength{\tabcolsep}{0.3em} % for the horizontal padding
% \centering
% \def\arraystretch{1.0}
% \begin{tabular}{lll}
% $\bullet$ 3x3 max pooling     & $\bullet$ 3x3 avg pooling  & $\bullet$ skip connection     \\
% $\bullet$ 3x3 separable conv  & $\bullet$ 5x5 separable conv & $\bullet$ 1x3 \& 3x1 conv \\
% $\bullet$ 3x3 dilated conv    & $\bullet$ 5x5 dilated conv \\
% \end{tabular}
% \end{table}
% \noindent We set the number of nodes in a cell as $B=4$.
%Therefore, the number of candidates in the search space of $\gO_{s}$ is $(1\times3\times6\times10\times{(6^2)}^{4})^{2} = 9.1\times10^{16}$; for $\gO_{l}$, the number of candidates for $\gO_{l}$ is $(1\times3\times6\times10\times{(8^2)}^{4})^2 = 9.1\times10^{18}$.
%The search space of $\gO_{l}$ is 100 times larger than the search space of $\gO_{s}$, while our experimental results show that the search cost on $\gO_{l}$ is only slightly higher than the search cost on $\gO_{s}$ by less than 10\%.
%Note that these two numbers are a roughly estimation

Once we obtain the topology structures of the normal cell and the reduction cell, we follow previous works to construct the overall CNN~\cite{Liu_2018_ECCV,liu2019darts,dong2019search}.
For CIFAR, the overall CNN is [image] $\rightarrow$ [N-Cell]$\times$N $\rightarrow$ [R-Cell] $\rightarrow$ [N-Cell]$\times$N $\rightarrow$ [R-Cell] $\rightarrow$ [N-Cell]$\times$N $\rightarrow$ [Softmax];
and for ImageNet, the overall CNN is [image] $\rightarrow$ [a pair of 3x3 Conv] $\rightarrow$ [3x3 Conv] $\rightarrow$ [N-Cell]$\times$N $\rightarrow$ [R-Cell] $\rightarrow$ [N-Cell]$\times$N $\rightarrow$ [R-Cell] $\rightarrow$ [N-Cell]$\times$N $\rightarrow$ [Softmax], where [N-Cell] and [R-Cell] indicate the normal and reduction cells, respectively.

\section{Methodology}
%In this section, we first introduce the template network in \Secref{sec:template-net} and the evaluator in \Secref{sec:network-estimator}.
%Then, the overall searching algorithm is introduced in \Secref{sec:searching}.
%Lastly, we briefly discuss our motivation and the connections with other NAS approaches in \Secref{sec:discussion}.
%(3) Preferably generate high-quality network candidates, evaluate them on the validation set, and then find the best one, introduced in \Secref{sec:evaluation}. We will introduce these steps in the following sections.

\subsection{Template Network}\label{sec:template-net}

The template network contains all candidate CNNs in the search space. The parameters of each candidate CNN are shared by a single template network.
It is non-trivial to make billions of candidate CNNs perform well after optimizing one template network.
To achieve this goal, we introduce a stochastic training strategy to optimize the template network by stochastically selecting the operations and inputs as below:
{\small
\begin{align}
    \mH_{i}^{c}   ~&~= f_{1}(\mI_{1}) + f_{2}(\mI_{2}),                   \label{eq:hybrid-net}  \\
    \mathrm{s.t.} ~&~\{\mI_{1}, \mI_{2}\} = \RMw(\gI_{i}^{c}, 2) \footnotemark, \label{eq:choice-I}         \\
                  ~&~\{f_{1}, f_{2}\} = \RMo(\gO, 2)             \footnotemark, \label{eq:choice-F}
\end{align}
}\noindent where\hspace{-3mm}
\footnotetext[2]{$\RMw(S,N)$ indicates a set of $N$ elements chosen randomly from set $S$ with replacement via a uniform distribution.}
\footnotetext[3]{$\RMo(S,N)$ indicates a set of $N$ elements chosen randomly from set $S$ via a uniform distribution, in the mean time, the index of a later sampled element should be not larger than the index of former element.
In implementation, we uniformly random sample an integer $r$ from [0, N(N+1)/2) and use \Eqref{eq:cal-r} to compute these two indexes.}, at the $i$-th node, we randomly sample two inputs $\mI_{1}$ and $\mI_{2}$ from the set $\gI_{i}^{c}$ with replacement; random sample two functions $f_{1}$ and $f_{2}$ from the set $\gO$ by restricting the index of $f_{1}$ in $\gO$ $\geq$ the index of $f_{2}$. This sample strategy can avoid the redundant candidates in the search space. For example, ($f_{1}$=3x3 conv,$f_{2}$=5x5 conv,$\mI_{1}$=$\mH^{c}_{1}$,$\mI_{2}$=$\mH^{c}_{1}$) is the same architecture as ($f_{1}$=5x5 conv,$f_{2}$=3x3 conv,$\mI_{1}$=$\mH^{c}_{1}$,$\mI_{2}$=$\mH^{c}_{1}$), but these two combinations are considered as different architectures during searching in some previous works~\cite{liu2019darts,pmlr-v80-pham18a,dong2019search}.

At each training iteration, the template network uniformly samples a candidate CNN, decided by \Eqref{eq:hybrid-net}$\sim$\Eqref{eq:choice-F}, and then it only optimizes template parameters of this sampled CNN.
This strategy allows us to optimize each candidate with equal opportunity, thus avoiding the Matthew effect. As a result, each candidate CNN is more likely to be fully trained compared to that in previous joint optimization strategies~\cite{liu2019darts,dong2019search,cai2019proxylessnas}.
We use ``the Matthew effect'' to refer that some quickly-converged candidates will get more chances to be further optimized in some NAS algorithms~\cite{liu2019darts,dong2019search,cai2019proxylessnas,pmlr-v80-pham18a}.
Besides, if we increase the cardinality of the function set $|\gO|$, the search space will grow exponentially, but the size of the template network will grow only linearly. This property allows us to search over large search space but only using a relatively small template network.
%
%\textbf{Note that}
%In experiments, we demonstrate that the proposed STS is superior to the trivial solution of optimizing TemplateNet.
%Here, we introduce two variants of STS, and will compare with them empirically later.
%(1) STS-LR: a stochastic approach with less randomness, in which the indexes of the sampled inputs and functions are the same for different cells.
%(2) Non-STS: directly training TemplateNet as the standard classification problem, in which $\mH_{i}^{c}$ is the sum of all possible function and input combinations.
%STS-LR and Non-STS are two potential solutions but worse than STS.

\subsection{Evaluator}\label{sec:network-evaluator}

We only optimize the template parameters on the training data, and a candidate CNN would thus be considered to generalize well if it can use learned template parameters to yield a low validation loss.
To find the best CNN, a trivial solution is traversing all candidates and evaluate them one by one, yet it would cost unaffordable computation time to cross over 10$^{16}$ candidates.
Some one-shot methods~\cite{bender2018understanding,brock2018smash} uniformly select a small amount of candidates to evaluate, where most uniformly selected candidates are useless.
To solve these issues, we design an evaluator to indicate the probability of each individual candidate CNN being likely to have a lower validation loss.
To represent this probability of each candidate, we encode one candidate CNN as a set of quadruples, and then define the probabilities over these quadruples.

\begin{algorithm}[t!]
\small
\caption{
The Searching Algorithm of {\NAME}
}
\label{alg:{\NAME}}
\small
\begin{algorithmic}
\Require        the whole available training data \\
\hspace{0.65cm} a template network with $\omega$ and an evaluator with $\alpha$
\State Split the whole available training data into the training set $\gD_{train}$ and the validation set $\gD_{val}$ for searching
\While{not converge}       \Comment{Optimize $\omega$ and $\alpha$}
  \State Sample training batch {\small $\gD_{t} = \{ (x_{i},y_{i})\}_{i=1}^{batch}$} from $\gD_{train}$
  \State Calculate {\small$\ell_{train} = \sum_{\gD_{t}}\ell(x_{i},y_{i})$} based on \Eqref{eq:hybrid-net}
  \State Update $\omega$ via gradients from the training loss {\small $\ell_{train}$}
  \State Sample validation batch {\small $\gD_{v} = \{ (x_{i},y_{i})\}_{i=1}^{batch}$} from $\gD_{val}$
  \State Calculate {\small$\ell_{val} = \sum_{\gD_{v}}\ell(x_{i},y_{i})$} based on \Eqref{eq:generator}
  \State Update $\alpha$ via gradients from the validation loss {\small $\ell_{val}$}
\EndWhile \\
After the above steps, we obtain the optimized $\omega$ and $\alpha$. \\
Initialize $\gA = \phi$ \Comment{Obtain Low-Validation-Loss Candidates}
\For{i=1; i $\leq$ $T$; i++}
  \State Sample an architecture $\va$ using \Eqref{eq:softmax-fg} to \Eqref{eq:assign-f}
  \State $\gA = \gA \supset \{\va\}$
\EndFor \\
Evaluate all candidates in $\gA$ with parameters extracted from $\omega$ \\
Select the candidate with the lowest validation loss
\Ensure the final selected candidate
\end{algorithmic}
  
\end{algorithm}

We introduce the encoding approach for the $i$-th node in the $c$-th cell, and one can easily infer the steps for other nodes. For simplicity, suppose we only search for one neural cell.
At first, we encode the choices of $\mI_{1}$ and $\mI_{2}$. Based on \Eqref{eq:choice-I}, there are $|\gI|$ choices for $\mI_{1}$ and $|\gI|$ choices for $\mI_{1}$ (we omit the subscript and superscript for simplicity). We thus use two vectors $\vf \in \gR^{|\gI|}$ and $\vg \in \gR^{|\gI|}$ to indicate the categorical choice for $\mI_{1}$ and $\mI_{2}$, and use its softmax-normalized value as the choice probability, which can be formulated as:
{\small
\begin{align}
    \hat{\vf} = \mathrm{softmax}(\vf) ~;~ \hat{\vg} = \mathrm{softmax}(\vg) ,   \label{eq:softmax-fg}   \\
    t \sim \gT(\hat{\vf}) \hspace{1mm}~;~\hspace{1mm} u \sim \gT(\hat{\vg}) , \label{eq:distribution_fg}\\
    \mI_{1} = \gI_{(t)}     \hspace{1mm}~;~\hspace{1mm} \mI_{2} = \gI_{(u)} ,
\end{align}
}\noindent where $\gT(\hat{\vf})$ represents the categorical distribution drawn by the vector $\vf$, so as $\gT(\hat{\vg})$.
$\mI_{1}$ and $\mI_{2}$ are chosen as the $t$-th and $u$-th element in $\gI$.
Similarly, there are $|\gO|(|\gO|+1)/2$ combination choices for $f_{1}$ and $f_{2}$, and therefore, we leverage a vector $\vh \in \gR^{|\gO|(|\gO|+1)/2}$ to indicate the categorical choice, which is formulated as:
{\small
\begin{align}
  \hat{\vh} = \mathrm{softmax}(\vh) ~\hspace{1mm}\rightarrow\hspace{1mm}~ r \sim \gT(\hat{\vh}) , \label{eq:distribution-h} \\
  r1 = \min\Big\{ n \in [~|\gI|~] : \sum_{k=1}^{n} k \geq r \Big\} ~;~ r2 = {r} - \sum_{k=1}^{r1\textrm{-}1} k \label{eq:cal-r}, \\
f_{1} = \gO_{(r1)} ~;~ f_{2} = \gO_{(r2)} ,\label{eq:assign-f}
\end{align}
}\noindent where $\gT(\hat{\vh})$ represents the categorical distribution drawn by the vector $\hat{\vh}$. $\gO_{(t)}$ and $\gO_{(u)}$ are the $t$-th and $u$-th functions in $\gO$, respectively. \Eqref{eq:cal-r} guarantees the indexes $r2 \leq r1$, which is consistent with \Eqref{eq:choice-F}.
\Eqref{eq:softmax-fg} to \Eqref{eq:assign-f} can sample one quadruple $(r1, r2, t, u)$ for one node based on the probabilities $\hat{\vf}$, $\hat{\vg}$, and $\hat{\vh}$, which are encoded by $\vf$, $\vg$, and $\vh$.
The set of quadruples for all nodes $\{(r1, r2, t, u)\}$ can represent one candidate architecture in the search space.

To enable the evaluator being able to indicate whether a candidate could result in a low validation loss, we need to optimize the parameters of this evaluator $\alpha=\{(\vh,\vf,\vg)\}$ on the validation set.
Since \Eqref{eq:softmax-fg} to \Eqref{eq:assign-f} are discrete, we use continues relaxation to calculate the output $\mH_{i}^{c}$ of a node, as follows:
{\small
\begin{align}
    \mH_{i}^{c} = \sum_{{r}=1}^{\frac{(|\gO||\gO|+1)}{2}} \hat{\vh}_{({r})}\nonumber \\
   \times (\sum_{t=1}^{|\gI|} \hat{\vf}_{(r1)}\gO_{(r1)} ({\gI}_{(t)}) + \sum_{u=1}^{|\gI|} \hat{\vg}_{(u)}\gO_{(r2)}(\gI_{(u)})) , \label{eq:generator} \\
    \mathrm{s.t.} \hspace{2mm} \text{calculate}~r1~\text{and}~r2~\text{as \Eqref{eq:cal-r}},
\end{align}
}
\noindent Based on \Eqref{eq:generator}, we can back-propagate gradients through the architecture encoding $\alpha$. To enforce the learned evaluator being able to reflect the validation loss of an architecture candidate, our objective for this evaluator is to minimize the validation loss.
Specifically, we forward the validation images through the template network with assistance of the evaluator via \Eqref{eq:generator}, and backward the validation loss\footnote{the validation loss could be a simple softmax with cross-entropy loss, and could also integrate other constrains about latency or memory size~\cite{cai2019proxylessnas}} to the evaluator's parameters.
In this way, after optimizing the evaluator, candidates sampled by the learned probabilities (\Eqref{eq:distribution_fg} and \Eqref{eq:distribution-h}) would be more likely to result in a high performance on the validation set.

\subsection{The Searching Algorithm of {\NAME}}\label{sec:searching}

%We have defined the TemplateNet and the SCG as above.
%In this section, we will introduce how to jointly train them in an end-to-end manner.
%Formally, we denote the parameters of the TemplateNet as $\sW$ and the parameters of SCG as $\sA = \{(\vf_{i}^{c}, \vg_{i}^{c}, \vh_{i}^{c}) | 1\leq i \leq B, 1 \leq c \leq M\}$, where $B$ is the number of nodes in a cell and $M$ is the number cells in a network. For both CIFAR and ImageNet, $M = 3~\times~N+2$.
We use $\omega$ to denote the parameters of the template network, i.e., the parameters of candidate functions in each node of each cell.
The searching algorithm of {\NAME} should (1) optimize parameters of the template network $\omega$ and parameters of the evaluator $\alpha$; (2) sample $T$=1000 low-validation-loss architecture candidates via the evaluator; and (3) evaluate these sampled candidates with the template parameters $\omega$ and choose the candidate with the lowest validation loss.

We show the overall searching algorithm in \Algref{alg:{\NAME}}, where $\ell(x,y)$ indicates the standard classification loss based on an input image $x$ with its label $y$.
We optimize $\alpha$ based on \Eqref{eq:generator} and $\omega$ based on \Eqref{eq:hybrid-net} in an alternative way.
The parameters of the template network are optimized on the training set, while the parameters of the evaluator are optimized on the validation set to guarantee the generalization ability of the searched model~\cite{liu2019darts}.

\textbf{Note that} we use different forward procedures for the template network and the evaluator: the template network uses \Eqref{eq:hybrid-net} and can enable each candidate perform well with the shared template network; the evaluator uses \Eqref{eq:generator} can help squeeze the candidate set for evaluation. 
After optimizing $\alpha$ and $\omega$, we sample $T$=1000 low-validation-loss candidates and select the one with the best one-shot performance.

\subsection{Connections with Other NAS Approaches}\label{sec:discussion}

Our proposed {\NAME} generalized over DARTS~\cite{liu2019darts} and one-shot NAS approaches~\cite{bender2018understanding,brock2018smash}.
DARTS directly can pick the best architecture, while this network capacity can not be correctly estimated in the validation set without fully training it. Besides, the architecture found by DARTS yields a performance with a high variance.
Comparatively, one-shot NAS can estimate the network performance correctly without additional training, while the architecture candidates are randomly picked with uneven quality instead of generating the ``best'' candidate~\cite{bender2018understanding,brock2018smash}.
Our {\NAME} is a new framework, which generalizes the above two typical streams and assimilates their benefits of accurate evaluation and high-quality architecture candidates selection.

To analyze the difference of technique details between our approach and others~\cite{liu2019darts}, we consider the following variants of our methods. \\
\textbf{{\NAME}}: our proposed search algorithm.\\
\textbf{{\NAME}-LR}: use a stochastic strategy to train the template network with less randomness, where the indexes of the sampled inputs/functions are the same for different cells.\\
\textbf{{\NAME}-NON}: optimize the template network without randomness, in which $\mH_{i}^{c}$ is the weighted sum of all possible function and input combinations as \Eqref{eq:generator}.\\
\textbf{{\NAME}-RAND}: randomly sample candidates for evaluation as previous one-shot approaches.\\
{\NAME}-NON is the same strategy as~\cite{liu2019darts}, and {\NAME}-RAND is the same strategy as~\cite{bender2018understanding,brock2018smash}.
We will show that {\NAME} is superior to {\NAME}-LR, {\NAME}-NON, and {\NAME}-RAND in experiments.

\begin{table*}[t!]
\centering
\setlength{\tabcolsep}{11pt}
\begin{tabular}{| l | c | c | c | c | c | c | c |} \hline
          Method                            & \makecell{GPU\\Days} &  M  &  C  & \makecell{Parameters} & \makecell{Error on \\CIFAR-10 (\%)}& \makecell{Error on\\CIFAR-100 (\%)} \\\hline
  DenseNet-BC~\cite{huang2017densely}       & $-$       & $-$ & $-$ &  25.6 MB      &          3.46             &    17.18  \\
  PyramidNet~\cite{han2017deep}             & $-$       & $-$ & $-$ &  26.0 MB      &          3.31             &    16.35  \\
      \hline\hline
 MetaQNN~\cite{baker2017designing}          & $>$80     & $-$ & $-$ &   11.2 MB     &          6.92             &    27.14  \\
 Net Transformation~\cite{cai2018efficient} & 10        & $-$ & $-$ &   19.7 MB     &          5.70             &     $-$   \\
 SMASH~\cite{brock2018smash}                & 1.5       & $-$ & $-$ &   16.0 MB     &          4.03             &     $-$   \\
 Hierarchical NAS~\cite{liu2018hierarchical}& 300       & 6   & 64  &   $-$         &          3.75             &    20.3   \\
 Progressive NAS~\cite{Liu_2018_ECCV}       & 150       & 11  & 48  &    3.2 MB       &          3.63             &     19.53  \\
 NASNet-A~\cite{Zoph_2018_CVPR}             & 2000      & 20  & 32  &    3.3 MB       &          3.41             &     19.70   \\
 AmoebaNet-A~\cite{real2019regularized}     & 3150      & 20  & 36  &    3.2 MB   &        3.34               &    $-$      \\
 ENAS~\cite{pmlr-v80-pham18a}               & 0.45      & 20  & 36  &    4.6 MB       &          3.54             &     19.43         \\
 NAONet~\cite{luo2018neural}                & 200       & 20  & 36  &    10.6 MB     &          3.18             &    $-$     \\
 \hline
 NASNet-A + CutOut~\cite{Zoph_2018_CVPR}    & 2000      & 20  & 32  &    3.3 MB    &   \textbf{2.65}           &  17.81$\dagger$     \\
PNAS + CutOut~\cite{Liu_2018_ECCV}          & 150       & 11  & 48  &    3.2 MB    &           $-$             &     17.63   \\
 DARTS + CutOut~\cite{liu2019darts}         & 4         & 20  & 36  &    3.4 MB    &          2.83             &    $-$     \\
 GHN + CutOut~\cite{zhang2019graph}         & 0.84      & 18  & 32  &    5.7 MB    &          2.84             &    $-$      \\
 ENAS + CutOut~\cite{pmlr-v80-pham18a}      & 0.45      & 20  & 36  &    4.6 MB    &          2.89             &  18.91$\dagger$   \\
 GDAS + CutOut~\cite{dong2019search}        & 0.84      & 20  & 36  &    3.4 MB    &          2.93             &  18.38     \\
    \hline
 {\NAME}-LR ($T$=1K) + CutOut               &   1.8      & 20  & 36  &  5.5 MB    &          2.81             &  17.88     \\
 {\NAME}-NON ($T$=1K) + CutOut              &   1.8      & 20  & 36  &  3.7 MB    &          3.12             &  18.27     \\
 {\NAME} ($T$=1) + CutOut                   &   1.7      & 20  & 36  &  4.5 MB    &          3.41             &  18.12     \\
 {\NAME} ($T$=1K)                           &   1.8      & 20  & 36  &  4.6 MB    &          3.56             &  19.38     \\
% {\NAME} ($T$=1K) + CutOut                  &   1.8      & 20  & 36  &  4.6          &          2.72             &  \textbf{17.25} \\\hline
 {\NAME} ($T$=1K) + CutOut                  &   1.8      & 20  & 36  &  4.6 MB    &          2.69             &  \textbf{17.25} \\\hline
\end{tabular}
\vspace{2mm}
\caption{
We compare {\NAME} and other algorithms on CIFAR-10 and CIFAR-100.
The top block presents state-of-the-art architectures designed by human experts.
The bottom block presents architectures that are automatically discovered by machine.
``$M$'' indicates the total number of cells in the CNN, and ``$C$'' denotes the number of the filter channel in the first cell.
``CutOut'' indicates the data argumentation approach~\cite{CUTOUT}.
$\dagger$ denotes the results reproduced by ourself.
The bottom five lines show results for different variants of our approach.
We run each model three times and report the mean error (lower is better).
}
\vspace{-2mm}
\label{table:CIFAR}
\end{table*}

\section{Experiments}\label{sec:experiments}

%We conduct experiments on three benchmarks, introduced in \Secref{sec:dataset} and show the experimental setup in \Secref{sec:setup}.
%We deliver the comprehensive empirical analysis in \Secref{sec:comparison} and \Secref{sec:ablation}.

\subsection{Experimental Setup}\label{sec:setup}

\textbf{Datasets.} CIFAR-10~\cite{krizhevsky2009learning} contains 60,000 images categorized into 10 classes. The training set has 5000 images per class, 50,000 images in total.
The test set contains 1000 images per class, 10,000 images in total.
{CIFAR-100}~\cite{krizhevsky2009learning} is similar to CIFAR-10.
It contains 50,000 training and 10,000 test images, categorized into 100 classes. All images are 32x32 colored ones.
{ImageNet}~\cite{russakovsky2015imagenet} is a large-scale image classification dataset, containing 1000 classes, 1.28 million training images and 50,000 validation images.
%. It has been a famous benchmark to validate the performance of different models. It contains one thousand classes, 1.28 million training images and 50,000 validation images.
%We use the same data argumentation as in \cite{xie2017aggregated,huang2017densely} for training. For evaluation, we only use the single-crop with input image size of $224^{2}$.

\textbf{Searching Setup.}
We search the normal CNN cell and the reduction CNN cell on CIFAR-10. For searching, the official training images are randomly split into the searching training set $\gD_{train}$ and the search validation set $\gD_{val}$ in \Algref{alg:{\NAME}}. $\gD_{train}$ contains 50\% of the official CIFAR-10 dataset, i.e., 25,000 images. $\gD_{val}$ contains the rest images.
The candidate function set $\gO$ has 6 different operations as introduced in \Secref{sec:background}.
The hyper-parameters to construct the whole CNN model are:
the number of nodes in a cell $B$, the initial channels of the first layer $C$, the number of repeated normal cells $N$.
By default, we set $C=16$, $B=4$, and $N=2$ to search the CNN cells.
Note that, the number of operations in $\gO_{s}$ is the same as ENAS~\cite{pmlr-v80-pham18a} and DARTS~\cite{liu2019darts}.

We train the template network and the evaluator with the batch size of 64 in 400 epochs.
To optimize the parameters $\omega$ of template network, we use the SGD optimization.
We start the learning rate of 0.025 and anneal it down to 0 following a cosine schedule. We use the momentum of 0.9 and the weight decay of 3e-4. We set the probability of path dropout as 0.1.
To optimize the parameters $\alpha$ of the evaluator, we use the Adam optimization~\cite{kingma2015adam} with the learning rate of 3e-3 and the weight decay of 1e-3.
To avoid the gradient explosion, we clip the gradient for both $\sW$ and $\sA$ by 10 during training.

%We use PyTorch~\cite{pytorch} for all experiments.
%For the small search space $\gO_{s}$,
\textbf{Computational Costs.}
Our {\NAME} would take about 40 hours to optimize both template network and evaluator on a single NVIDIA Tesla V100 GPU.
Evaluating $T$=1000 candidates costs less than three hours on a single GPU (about 13 seconds per candidate). Therefore, it requires about 43 GPU hours to obtain the final CNN structure.
%Re-training 100 networks from scratch requires about 85 hours on a single GPU. Therefore, it requires less than 6 GPU days to obtain the final network.
%Since one experiment costs nearly 43 GPU hours, we simply follow previous works~\cite{Liu_2018_ECCV,liu2019darts,Zoph_2018_CVPR} to set most hyper-parameters and only tune the total training epoch according to the validation set.
Note that when different NAS methods report their searching costs, they may use different hardware, e.g., NVIDIA GTX 1080Ti~\cite{liu2019darts} and NVIDIA P100~\cite{Zoph_2018_CVPR}.
We did not normalize the GPU cost of compared methods across different devices, and we use the numbers from their original papers.
%We will public codes and trained models upon acceptance.

\subsection{Compared with State-of-the-art Approaches}\label{sec:comparison}

{\bf Experiments on CIFAR.}
After we discover outstanding cells in the search space, we use the discovered topology with $C=36$ and $N=6$ to construct the CNN model for CIFAR-10 and CIFAR-100 following~\cite{liu2019darts,luo2018neural}.
We train this network by 600 epochs with the initial learning rate of 0.025.
We anneal the learning rate down to 0 with the cosine schedule.
The batch size is 96; the momentum is 0.9; and the weight decay is 5e-4.
The ratio of drop path is 0.2. Following~\cite{Liu_2018_ECCV,zoph2017NAS,Zoph_2018_CVPR,liu2019darts,dong2019search}, we use an auxiliary tower with the weight of 0.4 to train the network. We train each network three times and report the mean error.

{\bf Comparison with the state-of-the-art on CIFAR-10 and CIFAR-100.}
We compare the results of the found network with other state-of-the-art networks in \Tabref{table:CIFAR}.
``{\NAME} ($T$=1K)'' indicates the network found by our approach.
First, our {\NAME} is one of the most efficient algorithms, in which we complete the search procedure in 1.8 GPU days.
Second, among those efficient NAS approaches~\cite{zhang2019graph,pmlr-v80-pham18a,liu2019darts} (less than five GPU days), the network found by {\NAME} achieves the lowest error with similar or fewer parameters.
Other NAS approaches need more than 100 times computational costs than ours, whereas models found by most approaches have higher error with more parameters than our {\NAME}.
On CIFAR-100, our network achieves the best performance (a error of 17.25\%) among all compared methods.
On CIFAR-10, our network is slightly worse than NASNet-A (2.69\% test error vs. 2.65\% test error), however, NASNet-A needs more than 1000$\times$ computational costs than {\NAME}.

\begin{table*}[t!]
\centering
\setlength{\tabcolsep}{5pt}
\begin{tabular}{| c | l | c | c | c | c | c |} \hline
 &   Method  & \makecell{GPU days}          & Parameters & $+\times$ (million) & Top-1 Accuracy & Top-5 Accuracy \\ \hline
\multirow{4}{*}{\makecell{Human\\Expert}}
 & Inception-v1~\cite{szegedy2015going}             & $-$  &  6.6 MB  &      1448     &    69.8\%      & 89.9\%         \\
 & ResNet~\cite{he2016deep}                         & $-$  &  11.7 MB &      1814     &    69.8\%      & 89.1\%         \\
 & MobileNet-v2~\cite{sandler2018mobilenetv2}       & $-$  &  3.4 MB  &      300      &    72.0\%      & $-$ \\
 &  ShuffleNet~\cite{Zhang_2018_CVPR}               & $-$  &  $\sim$5 MB &      524      &    73.7\%      & $-$ \\ \hline\hline
\multirow{4}{*}{\makecell{NAS with\\more than\\100 GPU days}}
 & Progressive NAS~\cite{Liu_2018_ECCV}             & 150  & 5.1 MB   &    588        &    74.2\%      & 91.9\%         \\
 & NASNet-A~\cite{Zoph_2018_CVPR}                   & 2000 & 5.3 MB   &    564        &    74.0\%      & 91.6\%         \\
 & NASNet-B~\cite{Zoph_2018_CVPR}                   & 2000 & 5.3 MB   &    488        &    72.8\%      & 91.3\%         \\
 & NASNet-C~\cite{Zoph_2018_CVPR}                   & 2000 & 4.9 MB   &    558        &    72.5\%      & 91.0\%         \\
% & AmoebaNet~\cite{real2019regularized}             & AAAI19 & 3150 &  5.1 MB  &    555        &    74.5\%      & 92.0\%         \\
% & AmoebaNet~\cite{real2019regularized}             & 3150 &  5.3 MB  &    555        &    74.0      & 91.5         \\
% & AmoebaNet-C~\cite{real2019regularized}           & 3150 &  5.1 MB  &    535        &\textbf{75.1} & \textbf{92.1}\\
 \hline\hline
\multirow{8}{*}{\makecell{NAS with\\less than\\5 GPU days}}
 & DARTS~\cite{liu2019darts}                        & 4    & 4.9 MB   &    595        &    73.1\%      & 91.0\%  \\
 & GHN~\cite{zhang2019graph}                        & 0.84 & 6.1 MB   &    569        &    73.0\%      & 91.3\%  \\
 & SNAS~\cite{xie2019snas}                          & 1.5  & 4.3 MB   &    522        &    72.7\%      & 90.8\%  \\
 & GDAS~\cite{dong2019search}                       & 0.84 & 5.3 MB   &    581        &    74.0\%      & 91.5\%  \\ \cline{2-7}
 & {\NAME} (N=1 \& C=73)                            & 1.8  & 5.2 MB   &    597        &    73.3\%      & 91.4\%  \\
 & {\NAME} (N=2 \& C=58)                            & 1.8  & 5.3 MB   &    600        &    74.3\%      & 91.6\%  \\
 & {\NAME} (N=3 \& C=49)                            & 1.8  & 5.3 MB   &    584        &    74.1\%      & 91.9\%  \\
 & {\NAME} (N=4 \& C=44)                            & 1.8  & 5.4 MB   &    599        &    74.3\%      & 92.0\%  \\
\hline
\end{tabular}
\vspace{2mm}
\caption{
We compare networks found by {\NAME} and other approaches on ImageNet.
We report the model size, the computation cost, the top-1 accuracy, and the top-5 accuracy.
The top block shows the manually designed CNNs. The bottom two blocks indicate the automatically design CNNs.
$+\times$ indicates the number of multiply-add operations.
}
\vspace{-2mm}
\label{table:ImageNet}
\end{table*}

{\bf Comparison with other {\NAME} variants.}
In \Tabref{table:CIFAR}, we compare several variants of {\NAME} introduced in \Secref{sec:discussion}.
%``{\NAME}-LR ($T$=1K)'' and ``{\NAME}-NON ($T$=1K)'' indicate that we use STS-LR and Non-STS strategies to optimize SCG, respectively.
{\NAME}-NON is a straightforward approach to optimize the evaluator, however, the model found by {\NAME}-NON leads to the highest error.
{\NAME}-LR finds the model with more parameters but is inferior to the model found by {\NAME}.
In conclusion, the proposed {\NAME} is superior to its baselines, i.e., {\NAME}-LR and {\NAME}-NON.
``{\NAME} ($T$=1)'' indicates that we directly choose the best architecture from the evaluator via $\arg\max$ over $\vf$/$\vh$ as~\cite{liu2019darts}.
%In this approach, we directly use $\arg\max(\vh_{i}^{c})$, $\arg\max(\vf_{i}^{c})$, and $\arg\max(\vg_{i}^{c})$ to produce the set of quadruples as the final CNN model~\cite{liu2019darts}.
From \Tabref{table:CIFAR}, ``{\NAME} ($T$=1)'' finds a small CNN, however, this CNN yields a relatively higher error than our {\NAME}.
Therefore, the stochastic training strategy and the final candidate evaluation strategy are necessary to find a good architecture.

{\bf Scalability.}
We compare the search cost between the small space using the candidate function set $\gO$ and the large space using another candidate function set $\gO_{l}$.
This set $\gO_{l}$ adds two more functions into $\gO_{l}$: 3x3 dilated conv and 5x5 dilated conv.
It has eight functions in total, and its search space has about 9.1$\times$10$^{18}$ candidates.
With the large search space, training {\NAME} costs 50 GPU hours using the default hyper-parameters, and evaluating $T$=1K candidates takes less than three hours.
Therefore, even though the large search space is 100$\times$ larger than the small one, {\NAME} needs only about 18\% more GPU days to complete the search procedure.
Besides, the network found with $\gO_{l}$ achieves a similar performance compared to that of $\gO$. This shows that our {\NAME} can be successfully applied to much larger search space.
%Besides, the network found with $\gO_{l}$ achieves a error of 2.71\% on CIFAR-10 and a error of 17.31\% on CIFAR-100, which is similar to ``{\NAME} (T=1K)''.
%This result demonstrates the scalability of {\NAME}.

{\bf Experiments on ImageNet.}
We use the same cell structures found on the CIFAR-10 dataset to construct the CNN for ImageNet.
We adjust hyper-parameters $N$ and $C$ to make the network align with the ImageNet-mobile setting, i.e., under 600M FLOPs.
We train the network with a batch size of 256 over four GPUs in 250 epochs totally.
We warm-up at the first five epochs, start the learning rate with 0.1, and decrease it to 0 via the cosine scheduler~\cite{warmup2017}.
We set the momentum as 0.9 and the weight decay as 3e-5. Besides, the label smoothing is applied with a epsilon of 0.1.
An auxiliary tower with the weight of 0.4 is applied during training.

{\bf Comparison with the state-of-the-art on ImageNet.}
Since the training procedure of {\NAME} does not use any ImageNet images, this experiment can investigate the transfer-ability of the discovered network.
We use the same CNN structure found on CIFAR-10 with different $N$ and $C$ configurations.
These networks strictly matche the ImageNet-mobile setting.
We show the top-1 and top-5 accuracy in \Tabref{table:ImageNet}.
``{\NAME} (N=2 \& C=58)'' achieves a top-1 accuracy of 74.3\% on ImageNet.
Our network obtains competitive accuracy compared to efficient NAS approaches~\cite{liu2019darts,xie2019snas,zhang2019graph}.
%, our network is the best and achieves a higher accuracy than them.
AmoebaNet~\cite{real2019regularized} achieves a similar accuracy than ours, but it costs 3150 GPU days, which is 1750$\times$ more than ours.
In sum, our network is competitive to state-of-the-art networks, whereas {\NAME} needs acceptable search costs.

\subsection{Ablation Studies}\label{sec:ablation}

In \Secref{sec:comparison}, we deliver a brief comparison between {\NAME} and its variants w.r.t. the model size and the model accuracy.
In this section, we will give a more comprehensive analysis for different aspects of {\NAME}.

\begin{figure}[t!]
\begin{center}
\subfigure[{\NAME}]{
\label{subfig:sts-learn}
\includegraphics[width=0.47\columnwidth]{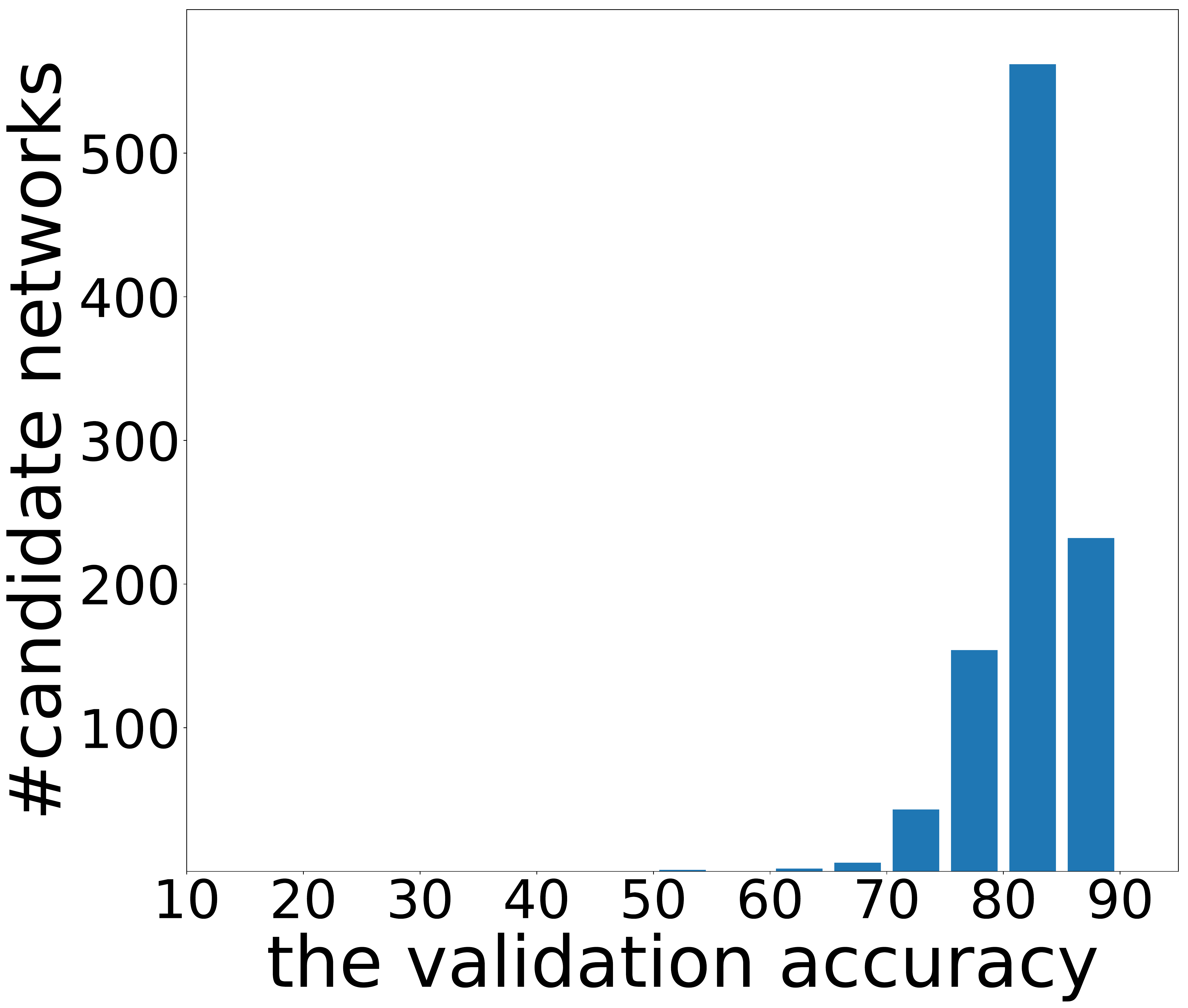}
}
\subfigure[{\NAME}-RAND]{
\label{subfig:sts-random}
\includegraphics[width=0.47\columnwidth]{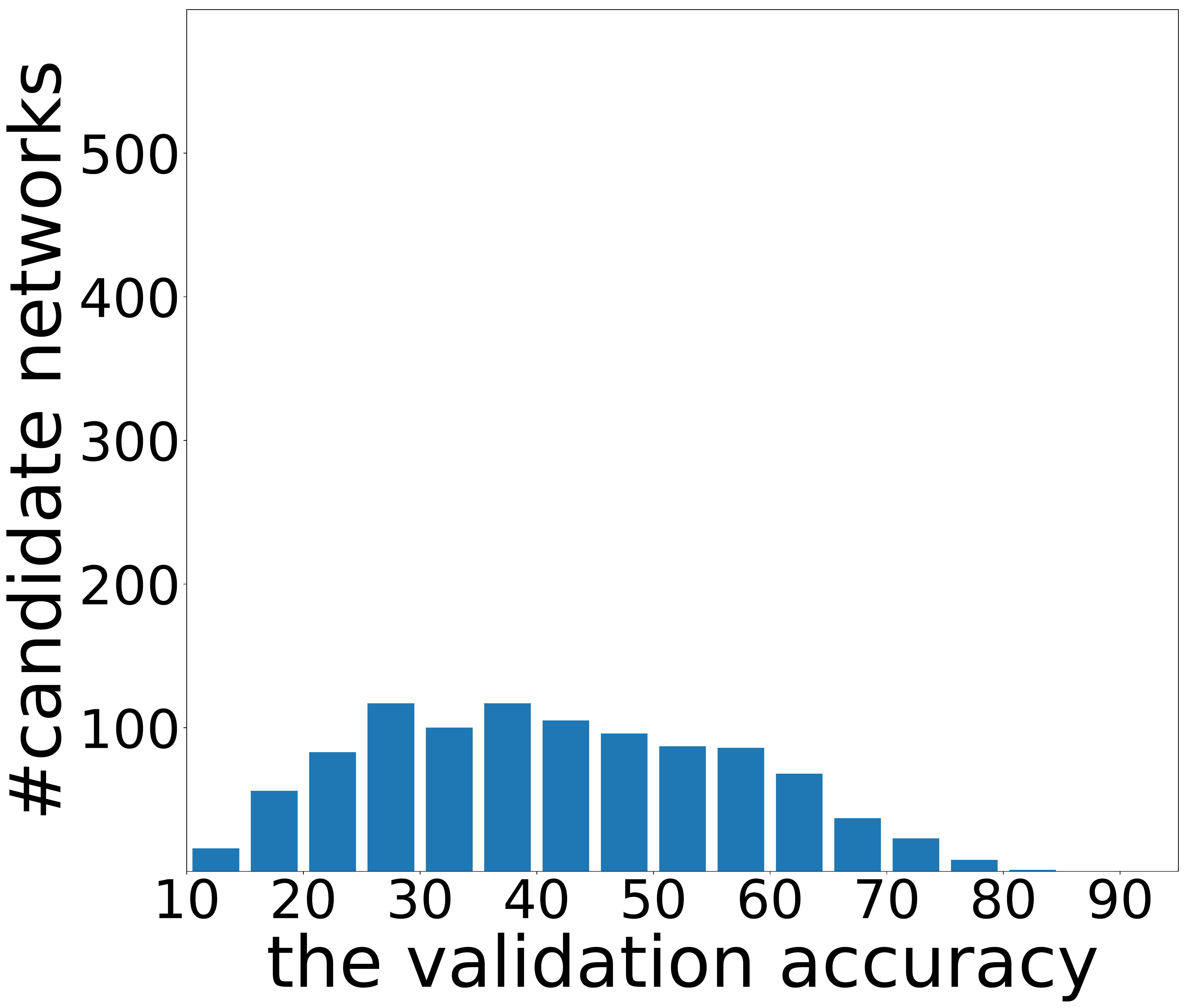}
}
\subfigure[{\NAME}-LR]{
\label{subfig:STS-LR}
\includegraphics[width=0.47\columnwidth]{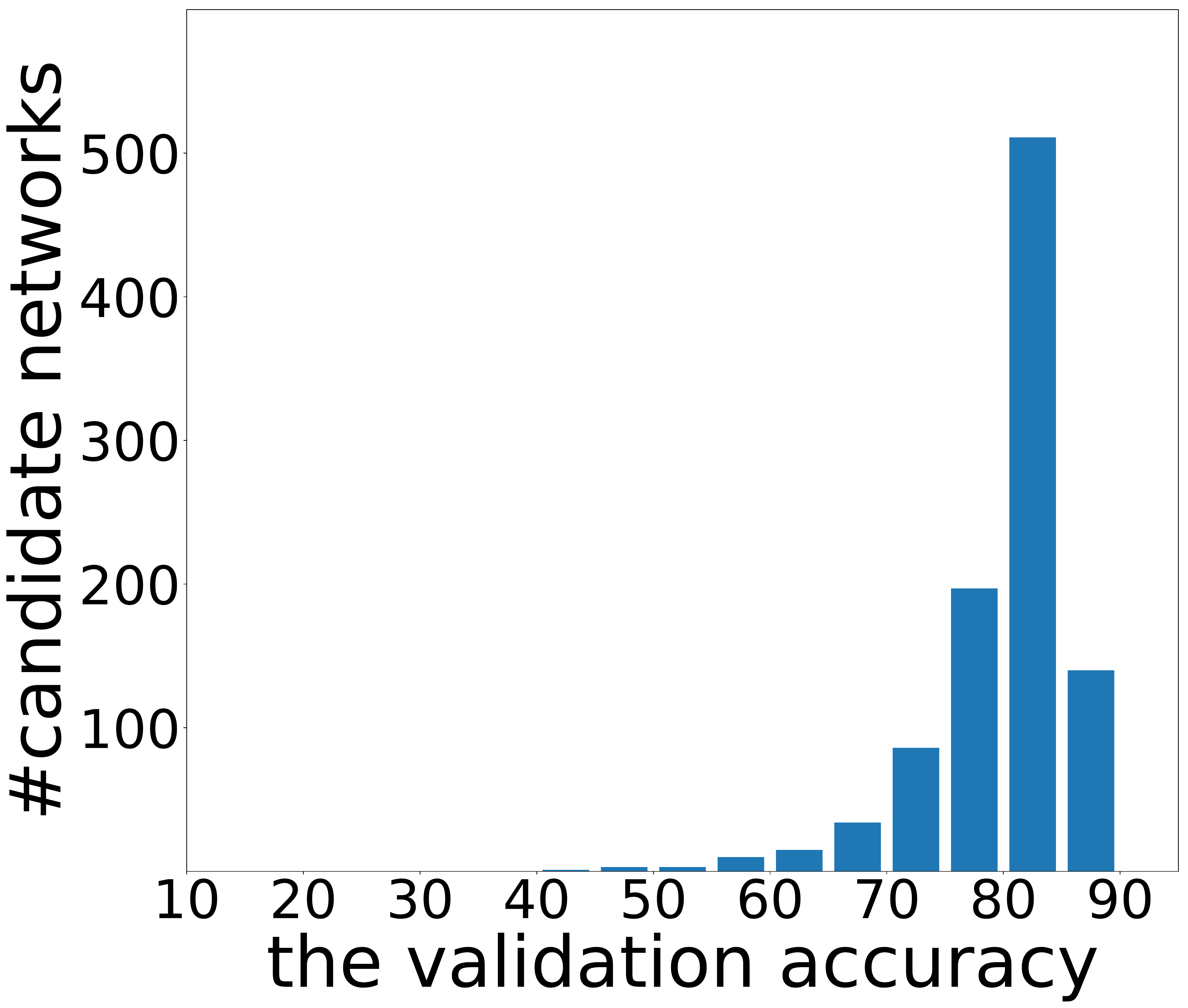}
}
\subfigure[{\NAME}-NON]{
\label{subfig:Non-STS}
\includegraphics[width=0.47\columnwidth]{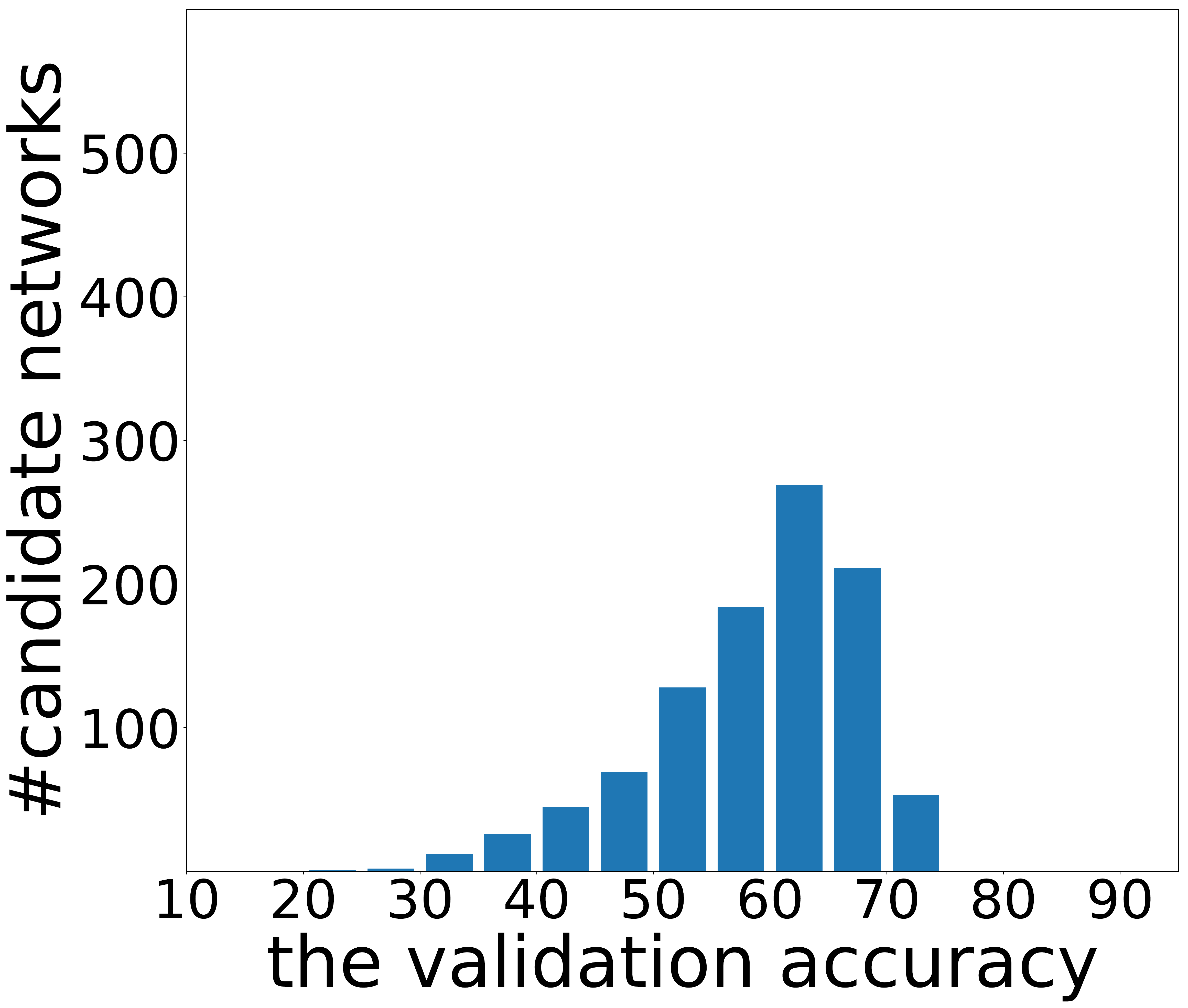}
}
\end{center}
\vspace{-2mm}
\caption[Captioning]{
Comparison of the candidate networks generated by different strategies.
\ref{subfig:sts-learn} shows the generated candidates from the proposed {\NAME}. \ref{subfig:STS-LR}, \ref{subfig:Non-STS}, and \ref{subfig:sts-random} show statistics of other three {\NAME} variants.
In each sub-figure, the x-axis and y-axis indicate the number of candidate networks and the validation accuracy, respectively.
}
\vspace{-3mm}
\label{fig:random-vs-learn}
\end{figure}

\textbf{The quality of estimated candidates.}
There are four options to generate candidate CNNs.
(a) the proposed {\NAME}. (b) {\NAME}-RAND~\cite{bender2018understanding}. (c) {\NAME}-LR, sharing the sampled indexes of operations and inputs for different cells (d) {\NAME}-NON, directly training the template network in the standard classification fashion as~\cite{liu2019darts}.
We use these four methods to generate 1000 candidate CNN and count their one-shot accuracy in a histogram.
In each sub-figure of \Figref{fig:random-vs-learn}, the x-axis indicates the one-shot validation accuracy and y-axis indicates the number of candidate CNNs.
Several conclusion can be made.
First, randomly generated candidates have much lower accuracy than all other compared strategies.
Second, {\NAME}-NON is better than the random approach but still inferior to {\NAME} and {\NAME}-LR.
Third, the performance of {\NAME} generated candidates is similar to that of {\NAME}-LR. However, taking a close look at the histogram, {\NAME} generate more accurate candidates than {\NAME}-LR, e.g., there are more candidates with the accuracy between 85\%$\sim$90\%.

%\textbf{The correlation between the accuracy using the template parameters and the retrained parameters.}
\textbf{Can the validation accuracy with template parameters reflect the ground truth validation accuracy?}
We want to investigate whether the validation accuracy using the template parameters can provide a robust relative ranking of different networks or not.
To achieve, we randomly sample 2000 networks, i.e., 1000 pairs.
We first evaluate these 2000 networks by using the template parameters on $\gD_{val}$.
We denote the obtained validation accuracy as \textit{acc}$_{t}$.
Then we retrain these 2000 networks from scratch from scratch by 100 epochs on $\gD_{train}$, and other hyper-parameters are the same as experiments in \Tabref{table:CIFAR}.
We evaluate these re-trained networks on $\gD_{val}$ and denote the accuracy as \textit{acc}$_{r}$, indicating the ground truth accuracy.
For each pair, if the comparison of \textit{acc}$_{t}$ is the same with the comparison of \textit{acc}$_{r}$, we count this pair is good; otherwise, we count this pair is bad.
Finally, we obtain more than 80\% pairs are good.
%
% This confirms that the template parameters can guarantee a strong relative ranking of networks.

The above two analysis paragraphs did not directly evaluate (1) how good the candidate sampled by the evaluator is and (2) how accurate the one-shot accuracy is.
To answer these two questions, we need a NAS dataset with ground truth accuracy of each candidate, where NAS-Bench-101~\cite{ying2019bench} is the only one.
However, our {\NAME} can not be directly evaluated on NAS-Bench-101~\cite{ying2019bench} due to its limitation.
We would investigate these open questions once a suitable NAS dataset being public.

\textbf{The effect of the number of generated candidates.}
In \Tabref{table:CIFAR}, we show that using $T$=1 finds a model with a higher error than using $T$=1K.
A small number of $T$, e.g., 10, will cause a high variance of the accuracy of discovered CNN.
If we increase the number of $T$, we could reduce the variance. In our experiments, $T$=1K is enough to discover a good network.
If we increase $T$ to 10K, we could potentially find better networks, but the search cost will be more than five GPU days.
Some approaches apply a progressive strategy to select the final CNN~\cite{bender2018understanding,zhang2019graph}.
For example, they first select the top 10 networks (ranked by the validation accuracy) and then retrain these networks with more epochs to get a precise validation accuracy.
These strategies are able to further reduce the performance variance of the discovered CNN.

\begin{figure}[t!]
\begin{center}
\includegraphics[width=\linewidth]{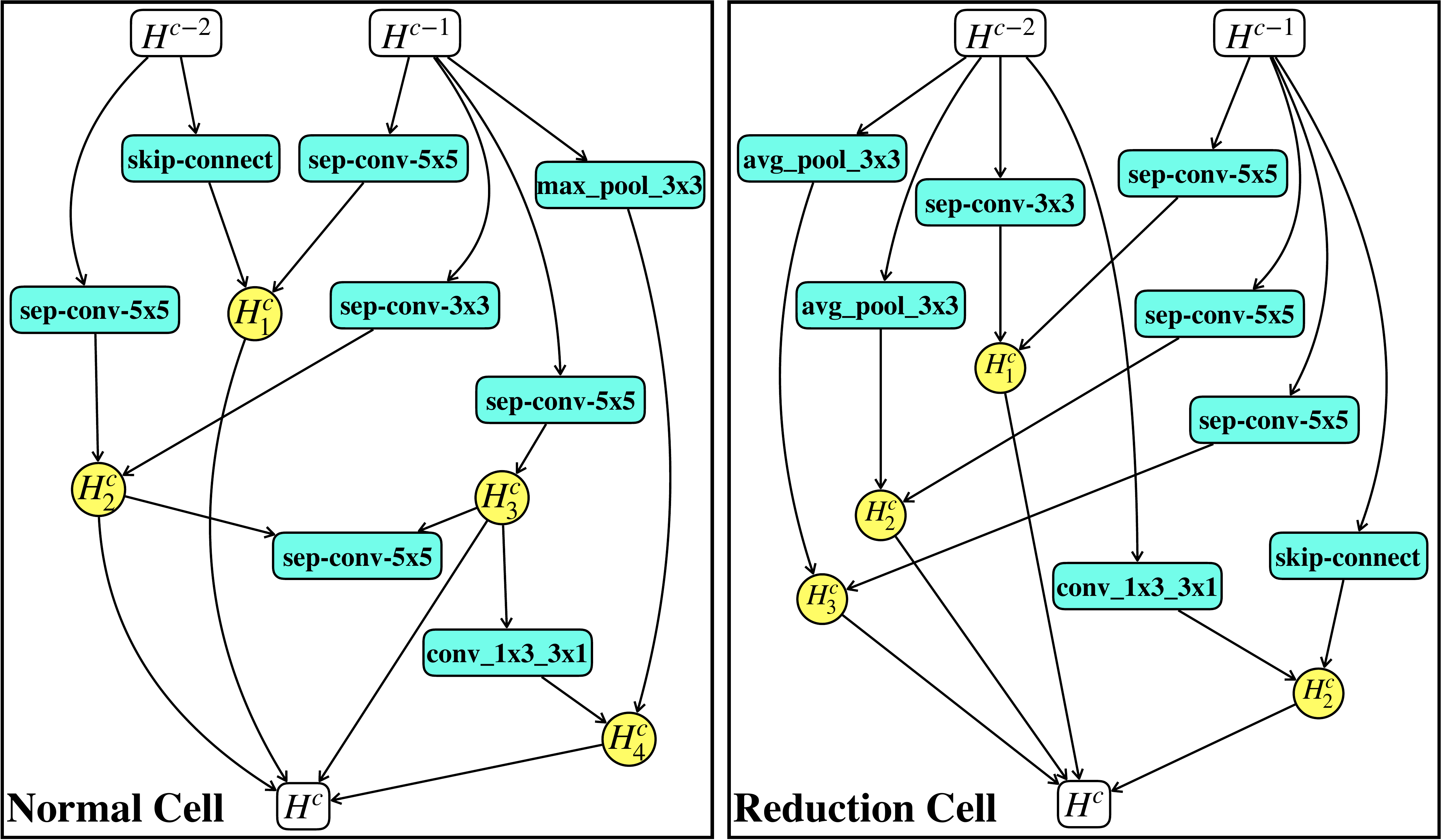}
\end{center}
\caption[Captioning]{
The left figure is the normal CNN cell that {\NAME} discovered on CIFAR-10.
The right figure is the reduction CNN cell that {\NAME} discovered on CIFAR-10.
}
\vspace{-2mm}
\label{fig:visualization}
\end{figure}

\textbf{Visualization.}
We visualize the discovered cells in \Figref{fig:visualization}.
Compared to manually designed cells~\cite{he2016deep,szegedy2016rethinking,xie2017aggregated}, the automatically discovered cells are much more complex and difficult to be designed by human experts.
Given the superior performance of NAS-discovered networks, it is necessary to devote more effort on this topic.

\textbf{Discussions.}
One limitation of {\NAME} is that when the search space is very large, e.g., 10$^{30}$ candidates, a small number of $T$ may not be able to find a good model. In this case, we have to increase the number of $T$, which will also increase the corresponding evaluation cost.
A more efficient training strategy for the template network and the evaluator can alleviate this problem.
We leave such interesting extensions for future work.

\section{Conclusion}

We propose the self-evaluated template network ({\NAME}) to search for the CNN with higher accuracy.
Compared to previous one-shot NAS approaches, {\NAME} significantly improves the quality of architecture candidates for the one-shot evaluation procedure. In this way, the sampled candidates of {\NAME} can cover better architectures, and thus can finally find an architecture with higher performance.
In experiments, {\NAME} can complete the search procedure within two GPU days. It finds a good CNN from more than 10$^{16}$ network possibilities, and this CNN achieves state-of-the-art performance on three benchmarks.

{\small
\bibliographystyle{ieee_fullname}
\bibliography{egbib}
}

\end{document}